\documentclass[10pt,twocolumn,letterpaper]{article}

\usepackage{iccv}
\usepackage{times}
\usepackage{epsfig}
\usepackage{graphicx}
\usepackage{amsmath}
\usepackage{amssymb}
\usepackage{multirow}
\usepackage{makecell}
\usepackage{bbding}
\usepackage[linesnumbered,ruled]{algorithm2e}

\usepackage[breaklinks=true,bookmarks=false]{hyperref}

\iccvfinalcopy 

\def\iccvPaperID{7320} 

\ificcvfinal\fi

\begin{document}

\title{Homogeneous Architecture Augmentation for Neural Predictor}

\makeatletter
\newcommand{\printfnsymbol}[1]{%
	\textsuperscript{\@fnsymbol{#1}}%
}
\makeatother
\author{Yuqiao Liu\thanks{Equal contribution.}~$^{1}$, Yehui Tang\printfnsymbol{1}$^{2}$, Yanan Sun\thanks{Corresponding author.}~$^{1}$\\
	\normalsize$^1$ College of Computer Science, Sichuan University. \\
	\normalsize$^2$  Key Lab of Machine Perception (MOE), Dept. of Machine Intelligence, Peking University. \\
	\small\texttt{lyqguitar@gmail.com; yhtang@pku.edu.cn;ysun@scu.edu.cn}\\
}

\maketitle

\ificcvfinal\thispagestyle{empty}\fi

\begin{abstract}
   Neural Architecture Search (NAS) can automatically design well-performed architectures of Deep Neural Networks (DNNs) for the tasks at hand. However, one bottleneck of NAS is the prohibitively computational cost largely due to the expensive performance evaluation. The neural predictors can directly estimate the performance without any training of the DNNs to be evaluated, thus have drawn increasing attention from researchers. Despite their popularity, they also suffer a severe limitation: the shortage of annotated DNN architectures for effectively training the neural predictors. In this paper, we proposed Homogeneous Architecture Augmentation for Neural Predictor (HAAP) of DNN architectures to address the issue aforementioned. Specifically, a homogeneous architecture augmentation algorithm is proposed in HAAP to generate sufficient training data taking the use of homogeneous representation. Furthermore, the one-hot encoding strategy is introduced into HAAP to make the representation of DNN architectures more effective. The experiments have been conducted on both NAS-Benchmark-101 and NAS-Bench-201 dataset. The experimental results demonstrate that the proposed HAAP algorithm outperforms the state of the arts compared, yet with much less training data. In addition, the ablation studies on both benchmark datasets have also shown the universality of the homogeneous architecture augmentation.  Our code has been made available at \url{https://github.com/lyq998/HAAP}.
\end{abstract}

\section{Introduction}
\label{sec_introduction}
Deep Neural Networks (DNNs) have been successfully applied to various challenging real-world problems, including image classification~\cite{he2016deep}, natural language processing~\cite{devlin2018bert}, speech recognition~\cite{zhang2017very}, to name a few. Well-designed architectures of DNNs are generally viewed as the deciding factor that the DNNs can achieve promising performance. Traditionally, designing architecture is nevertheless time-consuming, and is often exerted by experts with rich knowledge in both the DNNs and the task domain. Neural Architecture Search (NAS) is an automatic way to design the architectures of DNNs without such kind of expertise.

Generally, the NAS algorithm is composed of three different parts: search space, search strategy and performance estimation~\cite{elsken2018neural}. Particularly, given a search space predefined, the search strategy should find a well-performed architecture when the search terminates. In literature, there are mainly three techniques used as the search strategies: reinforcement learning~\cite{sutton2018reinforcement}, gradient-based algorithms~\cite{lecun1988theoretical} and Evolutionary Computation (EC)~\cite{back1997handbook}. However, no matter which technique is used, all NAS algorithms need to estimate the performance of the DNNs searched in order for the effective proceeding of search. In most NAS algorithms, the search strategy evaluates the solutions individually, which inevitably gives rise to heavy computational overheads. For instance, on the commonly used CIFAR-10 benchmark dataset~\cite{krizhevsky2009learning}, the Large-Scale evolution NAS algorithm~\cite{real2017large} used 250 GPU computers for 11 days. Not only that, the NAS algorithm~\cite{zoph2016neural} used 800 GPUs for nearly one month. Unfortunately, this is unaffordable for most academic researchers. To address this issue, various algorithms have been designed to speed up the estimation process without numerous computation resource, which can be classified into four different categories: weight inheritance~\cite{real2017large}, early stopping policy~\cite{sun2018particle, fujino2017deep}, reduced training set~\cite{sapra2020constrained} and neural predictor~\cite{tang2020semi}.

The neural predictor requires a number of well-trained architectures, and then a regression model is trained to map the architectures and the corresponding performance values. When a new DNN is generated, its performance is directly predicted by the regression model. Compared to other speed-up algorithms mentioned above, the neural predictors can provide satisfactory prediction result, thus is popular among the community.

Many previous works focused on designing better regression models to improve the performance of predictors~\cite{sun2019surrogate, wen2020neural}. They try to achieve a better fit from the perspective of the regression model. However, the big issue in neural predictor is actually in the training dataset. In practice, it is prohibitively unaffordable to obtain a large training dataset. Even with the ultra high performance hardware, for example, it will take about 32 minutes to train a neural network on the TPU v2 accelerator~\cite{ying2019bench}, which means only nearly 45 annotated training data can be obtained one day. Generally, neural predictor is trained with a small dataset of annotated architectures, e.g., Wen \textit{et al.}~\cite{wen2020neural} only used 119 annotated architectures to build the neural predictor in practice. The main reason for the poor performance of the neural predictor is not the regression model, but the limited training data. As a result, how to make full use of the existing limited data without increasing the computational cost is an important issue in neural predictor.

In this paper, we propose a novel data augmentation strategy in the space of neural architectures by exploring their \textit{Homogeneous} properties. The \textit{Homogeneous} augmentations eliminate the influence of layer order, making the neural predictor pay more attention to the overall layer type.
Specifically, the architecture augmentation works by swapping the inner orders to generate a group of homogeneous representations. To effectively represent the intrinsic properties of architectures, one-hot encoding strategy is also developed. The flowchart of the proposed Homogeneous Architecture Augmentation for Neural Predictor (HAAP) is shown in Fig.~\ref{fig_flowchart}. The experiments on NAS-Bench-101~\cite{ying2019bench} and NAS-Bench-201~\cite{dong2020bench} demonstrate the effectiveness of the proposed architecture augmentation method, and show the superiority of HAAP compared with the state of the arts. Not only that, the \textit{Homogeneous} augmentations can also be combined with the state-of-the-art neural predictors to improve their predictions.

The reminder of this paper is organized as follows. The related works are provided in Sec.~\ref{sec_background}. Sec.~\ref{sec_algorithm} gives the implementation details of the proposed approach. Experiments and extensive experimental results demonstrate the effectiveness of HAAP in Sec.~\ref{sec_experiment}. Finally, Sec.~\ref{sec_conclusion} is for the conclusion and future works.

\section{Related Works}
\label{sec_background}
In this section, we will first introduce the encoding strategy of NAS algorithms, which is the base of the proposed algorithm. Immediately after, the state-of-the-art algorithms are reviewed and their limitations are summarized to justify the necessity of the proposed algorithm.

\subsection{Encoding Strategy of NAS Algorithms}
\renewcommand{\thefootnote}{\arabic{footnote}}
The encoding space (\textit{i.e.}, search space)
in NAS can be divided into four different categories based on the basic unit searched: the layer-based, the block-based, the cell-based and the topology-based.

The basic units in the layer-based encoding space are the primitive layers of DNNs. The layer-based encoding space is often large and building architecture in this space acquires more details and information, resulting in the search strategy putting many efforts. EvoCNN~\cite{sun2019evolving} is a typical example of layer-based space. In contrast, the basic units in the block-based encoding space are the blocks which are the combinations of the primitive layers, such as ResNet block and DenseNet block used by AE-CNN~\cite{sun2019completely}. With the help of the integrated blocks, the architectures searched from the block-based encoding space are more likely to be well-designed with fewer parameters. One special case in the block-based encoding space is that all the blocks are all the same, which generates the cell-based encoding space. The architectures found in the cell-based encoding space are made of stacked cells. Since all the cells are the same, the encoding information of one cell can represent a corresponding architecture by stacking the same cell multiple times. The well known AmoebaNet-A~\cite{real2019regularized} is an example of cell-based space. The topology-based encoding space mainly concentrates on the connections of the units rather than their internal structures. 

For cell-based encoding space, there are multiple ways to encode the connections of layers in cells. The adjacency matrix is a common way to represent the connections, where ``$1$'' means the two layers are connected while ``$0$'' means the opposite. This representation method has been widely adopted as the base encoding strategy by existing NAS algorithms. For example, Genetic CNN~\cite{xie2017genetic} used a variant of the adjacency matrix where the unimportant elements in the matrix are pruned and the remaining elements are flattened into a binary vector.

\begin{figure*}
	\centering
	\includegraphics[width=1.0\linewidth]{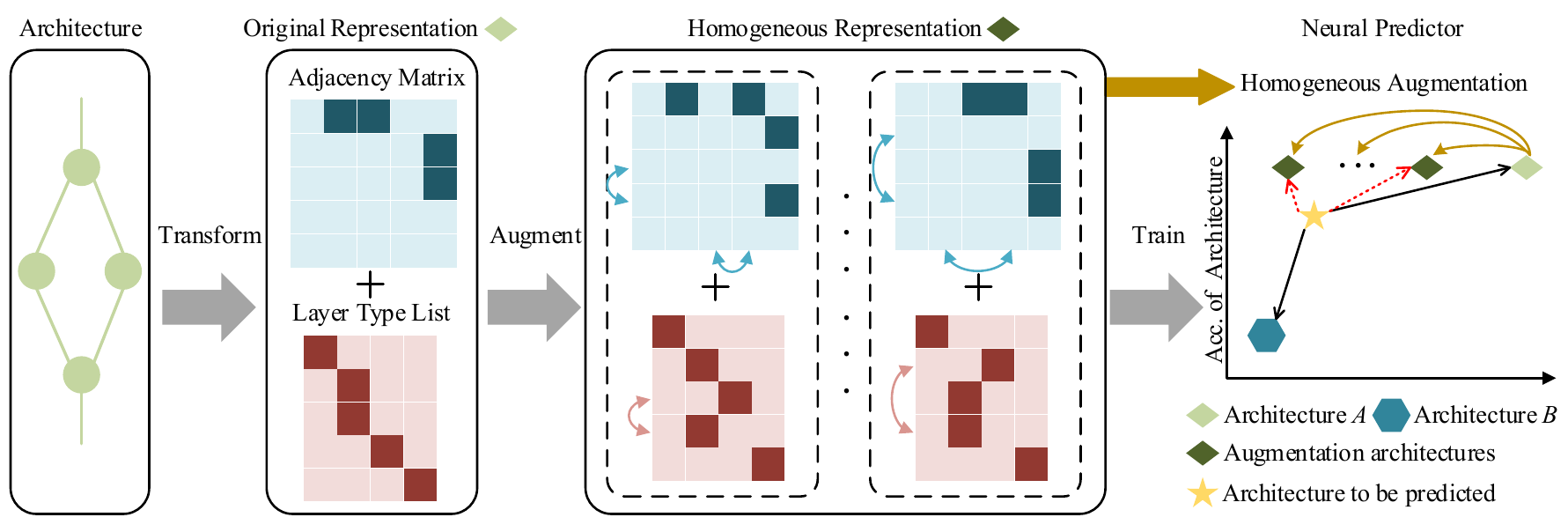}
	\caption{The flowchart of the proposed HAAP. Architectures in different encoding spaces are first transformed into the representation of adjacency matrix and layer type list where one-hot encoding strategy is used. Please note that, we only display one neural architecture in the figure for convenience. Then the original representation is augmented into its homogeneous forms. The original data and the augmented data are all used to train the neural predictor. If the architecture augmentation is not used, the architecture to be predicted will refer more to the accuracy of architecture $B$. When the architecture augmentation is used, the architecture to be predicted will be more inclined to refer to the accuracy of the augmentation architectures, and then get a more accurate prediction.}
	\label{fig_flowchart}
\end{figure*}

\subsection{Neural Predictor}
\label{sec_predictor}
There are five representative state-of-the-art neural predictors: Peephole~\cite{deng2017peephole}, E2EPP~\cite{sun2019surrogate}, Semi-Supervised Assessor of Neural Architectures (SSANA)~\cite{tang2020semi}, Neural Predictor for Neural Architecture
Search (NPNAS)~\cite{wen2020neural} and ReNAS~\cite{xu2021renas}. All of them have been investigated on image classification tasks owning to various well-designed architectures and available benchmark datasets for classifying images. Recently, we also notice that some public architecture datasets have been made available~\cite{ying2019bench,dong2020bench}. This will make the prediction more efficient and the comparison fairer with providing more training data. 

As mentioned in Sec.~\ref{sec_introduction}, most of the neural predictors mainly aim at the design of the regression models. For example, Peephole and NPNAS used different efficient regression models for better prediction, while ReNAS focused on the redesign of training loss of the model. In this paper, we demonstrate that without a complicated regression model, the classical regression models (e.g. random forest) can also make an accurate prediction with the utilization of the proposed architecture augmentation method.

\section{Approach}
\label{sec_algorithm}
Building the neural predictor can be mathematically represented as a regression task. Supposing the predictor is represented by a regression model $R$. After $R$ has been trained with the training data $[\textbf{X},\textbf{y}]$, where $\textbf{X}=\{ X_{1}, X_{2}, \dots, X_{N}\}$ denotes the encoding of architecture information and $\textbf{y}=\{ y_{1}, y_{2}, \dots, y_{N} \}$ denotes the corresponding performance value of $\{ X_{1}, X_{2}, \dots, X_{N} \}$, the prediction $\hat{\textbf{y}}$ of test data $\textbf{X}^{test}$ can be predicted by $R$. Particularly, the training process of $R$ can be described by Equation~\ref{equ_problem}:

\begin{equation}
\label{equ_problem}
\min_{T_{p}}\frac{1}{N}\sum_{n=1}^{N} \mathcal{L}(R(T_{p}, X_{n}), y_{n})
\end{equation}
where $T_{p}$ is the trainable parameters of the regression model $R$, and $\mathcal{L}(\cdot)$ denotes the loss function of $R$. We choose the $\ell_2$ norm as the loss function, which is widely used in neural predictors~\cite{wen2020neural, tang2020semi}.

\subsection{Architecture Encoding}
In principle, most of the neural architectures can be represented by a standard Directed Acyclic Graph (DAG), where the vertices represent the layers and the edges represent the connections. Therefore, each $X_{n}$ in $\textbf{X}$ can be represented by the layer type list $x^{t}_{n} = \{ x^{t}_{n1}, x^{t}_{n2}, \dots, x^{t}_{nN_{l}} \}$ and the corresponding adjacency matrix $x^{m}_{n}$, where $N_{l}$ denotes the number of layers. Although some architectures in other encoding spaces are not represented by standard DAG, where the vertices represent the connections while the edges represent the layers, we can use a transformation to change the form into the standard type. For example, to transform the architectures in NAS-Bench-201, we can firstly give a fixed order for edges, and then reduce the useless layers. The detailed transformation method is provided in supplementary materials.

The traditional hard encoding~\cite{xu2021renas} uses an integer vector to encode layer type $x_{n}^{t}$ directly. Then, the type vector is broadcasted into the adjacency matrix $x^{m}_{n}$ which can be expressed in Equation~\ref{equ_traditional_xn}:
\begin{equation}
\label{equ_traditional_xn}
X_{n} = x^{m}_{n} \times strech(integer(x^{t}_{n}))
\end{equation}
where the $integer(\cdot)$ transforms verbal type list to integer vector, and $strech(\cdot)$ can strech a vector into a square matrix. As a result, the encoding $X_{n}$ is a sparse matrix which contains much redundant dimensions. In addition, the relationships and the Euclidean distances between layers are misrepresented by integers. To this end, we come up with a binary encoding strategy which takes the use of the one-hot encoding for type list. 

\begin{figure}
	\centering
	\includegraphics[width=1.0\linewidth]{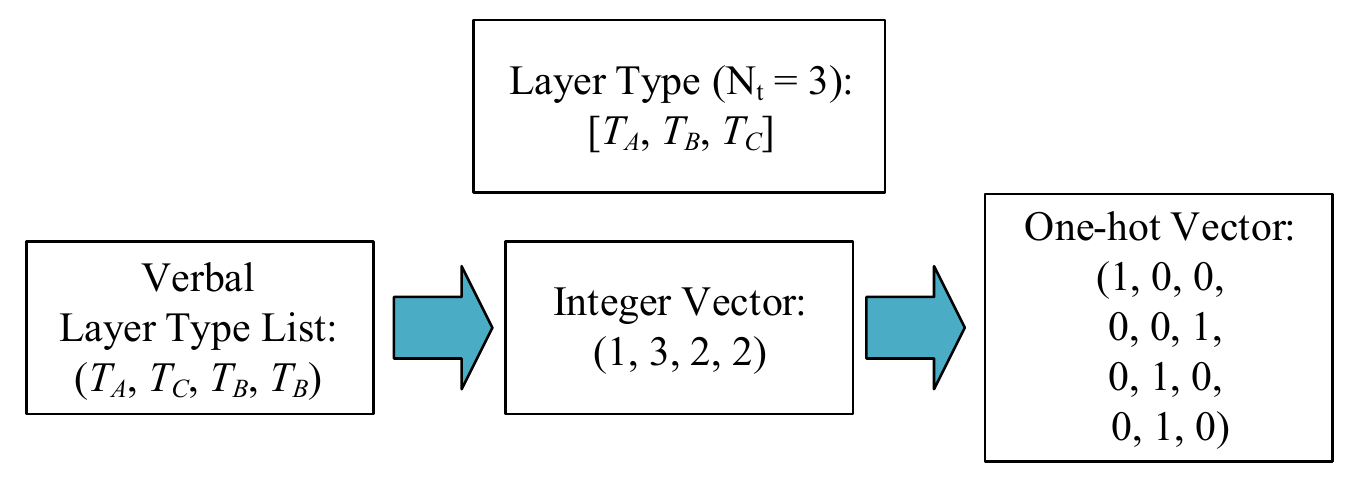}
	\caption{One-hot encoding process for the layer type list.}
	\label{fig_one_hot}
\end{figure}

The encoding strategy used in this paper is easy to implement. Equation~\ref{equ_xn} shows how to convert the original $x_{n}^{m}$ and $x_{n}^{t}$ into a binary vector.
\begin{equation}
\label{equ_xn}
X_{n} = concat\{flattened(x^{m}_{n}) , one\mbox{-}hot(x^{t}_{n})\}
\end{equation}
Specifically, the two-dimensional adjacency matrix $x^{m}_{n}$ is flattened (with $flattened(\cdot)$) into an one-dimensional vector by row. Meanwhile, the layer type list is transformed into the one-hot encoding (with $one\mbox{-}hot(\cdot)$). Finally, these two one-dimensional vectors are concatenated into $X_{n}$ (with $concat(\cdot)$). Fig.~\ref{fig_one_hot} provides an example to explain the one-hot encoding strategy operating on the layer type list, where the number of types $N_{t}$ is 3, and $T_{A}$, $T_{B}$ and $T_{C}$ denote the three types of layers in the list. To achieve the encoding, the first is to transfer the verbal expression into an integer vector. The second is to employ the one-hot encoding to replace the integers accordingly. The one-hot encoding uses multi dimensional space to represent the layer type to ensure the Euclidean distances between different layer types are the same, which is positive to the regression models.

\subsection{Homogeneous Architecture Augmentation}
\begin{figure}
	\centering
	\includegraphics[width=1.0\linewidth]{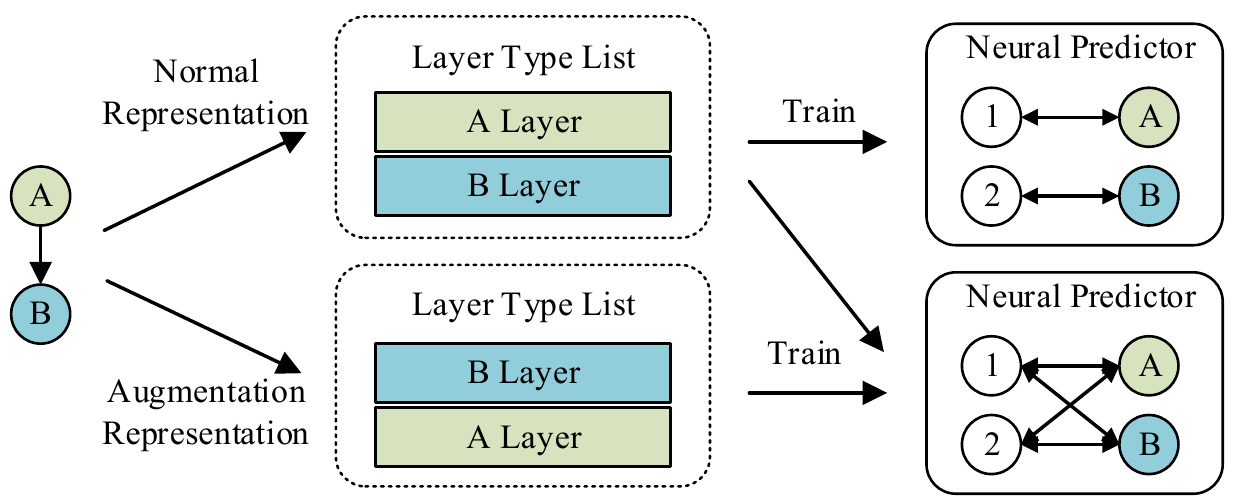}
	\caption{An example to illustrate how the homogeneous augmentation can better train the neural predictor.}
	\label{fig_order}
\end{figure}
Generally, a common DNN architecture begins with an input layer and ends with an output layer~\cite{ying2019bench,sun2019completely}. In this paper, we use \textit{In} and \textit{Out} layers to denote both two layers separately. Please note that the proposed architecture augmentation method is suitable for the case that the neural network has sole input and sole output. This is also the most common case of DNNs. For example, the image classification task only requires one image as input and the neural network outputs the predicted category.

As introduced above, the adjacency matrix and the layer type list make up the neural architecture, where the adjacency matrix describes the topological relationship of the layer type list. Therefore, there should be no order in the layer type list. However, the layers close to the \textit{In} layer are always placed in front of the layer type list, and the layers close to the \textit{Out} layer can only be placed at the back. In addition, null layers are generally placed at the end~\cite{wen2020neural, xu2021renas}. In this way, the layers that should have no order and position are sorted in a breadth-first traversal manner. This sorting can mislead the neural predictor and make the predictor pay attention to the relationship between layer type and the specified location which is meaningless. What we have to do is to eliminate the influence of layer position, making the neural predictor pay more attention to the overall layer type and some invariants in the augmentation of the neural architecture, such as the total number of different types of layers, instead of focusing on the type of layer in a specific location. Fig.~\ref{fig_order} is an example. Shown on the left is a part of a neural network, including two layers A and B. If the neural predictor is only trained with the normal representation of the architecture, the predictor will bind position 1 to A, and position 2 to B. However, if the predictor is trained with normal and augmented representation together, the influence of the layer order on the predictor will be weakened.

By observing this, the original representation can be augmented with the ``unordered" character of the layer type list. However, not all the layers in the layer type list are unordered. The \textit{In} and \textit{Out} layers belong to placeholder set and they are fixed at the first and last positions of the layer type list in all architectures. Thus, it is useless to change their positions. The other layers in the middle of the layer type list belong to operation set which principally determines the performance of an architecture. Actually, only these layers are unordered in layer type list. Assuming that the layer type list is $x^{t} = \{ x^{t}_{1}, x^{t}_{2}, \dots, x^{t}_{N_{l}} \}$, $x^{t}_{1}$ is the \textit{In} layer and $x^{t}_{N_{l}}$ is the \textit{Out} layer. Since for $\forall x^{t}_{j},\ j\in\{2,3,\dots,N_{l}-1\}$ could be in any position, therefore there are $(N_{l}-2)!$ possible permutations. We use these different permutations to augment the training dataset by a factor of $(N_{l}-2)!$.

However, if the layer type list is changed, the adjacency matrix has to make changes accordingly to make sure that the actual architectures of the representation are the same. This is because the order of rows and columns of the adjacency matrix is related to the order in the layer type list. Specifically, the \textit{i-th} row of the adjacency matrix denotes the output of the \textit{i-th} layer in the layer type list, and the \textit{j-th} column denotes the input of the \textit{j-th} layer. The ``Homogeneous Architectures" part of Fig.~\ref{fig_flowchart} shows an example to illustrate the architecture augmentation for neural predictors. In this example, we assume the number of layers $N_{l}$ equals to 5 and the neural architecture contains a null layer. The original representation of the architecture is on the top, and we term these different representation of architecture in dashed box as homogeneous representation, because they represent the same architecture in essence. Specifically, comparing the first homogeneous representation displayed in Fig.~\ref{fig_flowchart} with the original representation, the order of \textit{3-rd} layer and \textit{4-th} layer is swapped from the original one. To maintain the representation of architecture, the corresponding row and column in the adjacency matrix need to swap according to the swapping of the layer type list. Whether to swap rows or columns first, the matrix ends up with the same effect.

The general process of constructing more training data with the proposed architecture augmentation method is shown in Fig.~\ref{fig_flowchart}. Specifically, the input can be architectures in different encoding spaces, such as the binary vector used in Genetic CNN~\cite{xie2017genetic} or the operations represented by edges in NAS-Bench-201~\cite{dong2020bench}. The first step is to transform the different encoding strategies into the standard DAG form like NAS-Bench-101~\cite{ying2019bench}. Because of the character of the regression model, the encoded architecture information must be fixed length. We refer the padding approach in~\cite{xu2021renas} to achieve this goal, by adding zeros into the adjacency matrix and the layer type list. The position of starting to pad zeros is the penultimate element in the layer type list, while the starting positions in the adjacency matrix are the penultimate row and column. We define the zero in the layer type list as null layer type for convenience. The next step is the homogeneous architecture augmentation. Including the original representation, the augmentation ends up with $N_{c}$ architecture augmentations where $N_{c}=(N_{l}-2)!$ except for \textit{In} and \textit{Out} layers.
Next, the unnecessary dimensions in the adjacency matrix and the layer type list are reduced. To be more specific, the first element (\textit{In} layer) and the last element (\textit{Out} layer) in the layer type list, and the first column and the last row in the adjacency matrix will be reduced. Finally, the layer type list is converted to a binary vector and concatenate with flattened adjacency matrix.

The training dataset can be expanded by a factor of $N_{c}$ in the end. 
In principle, the generated data are reliable because of the homogeneous representation. It makes sense that all the homogeneous representation of $X_{n}$ correspond to the same performance value $y_{n}$. 
Similar to the image augmentation (\textit{e.g.}, flip, rotation, \textit{etc.}), the proposed augmentation can generate a more comprehensive dataset with more possible training points, which reduces the distribution gap between training data and unknown data~\cite{shorten2019survey}.
As a result, the regression model $R$, i.e. the neural predictor, can be trained better with more training data augmented.

\begin{figure}
	\centering
	\includegraphics[width=0.7\linewidth]{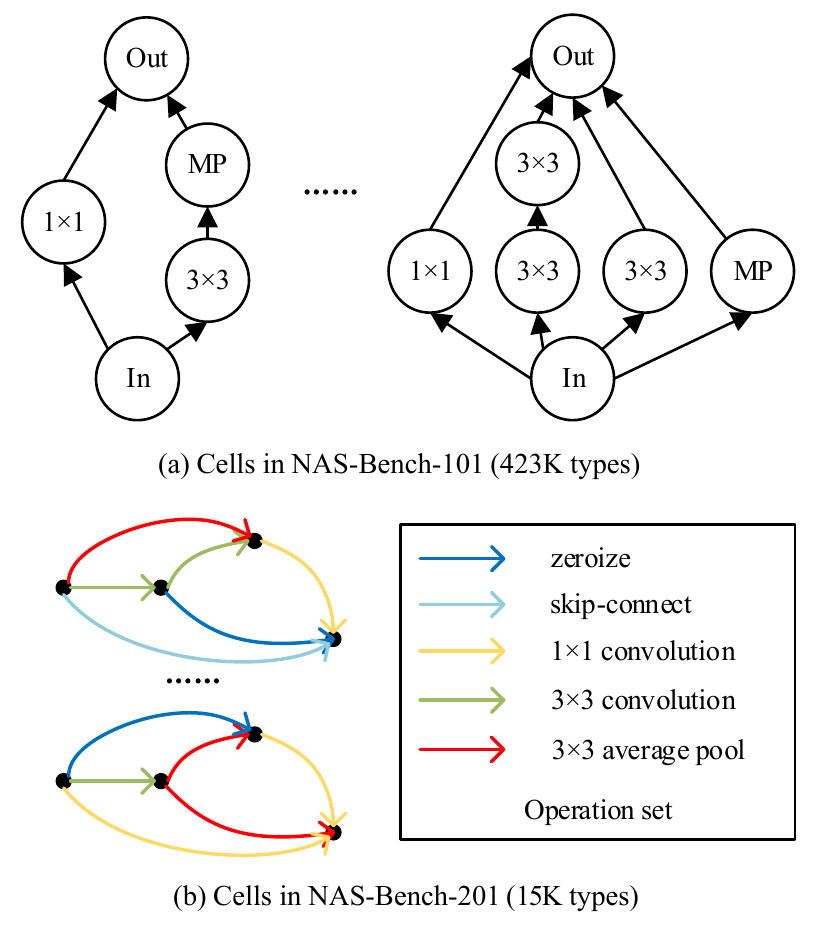}
	\caption{(a) Cells in NAS-Bench-101 where 1$\times$1 denotes 1$\times$1 convolution, 3$\times$3 denotes 3$\times$3 convolution and MP denotes 3$\times$3 max pooling. (b) Cells in NAS-Bench-201.}
	\label{fig_nas_bench}
\end{figure}

\section{Experiments}
\label{sec_experiment}
We choose the NAS-Bench-101~\cite{ying2019bench} and the NAS-Bench-201~\cite{dong2020bench} as the experimental datasets. The NAS-Bench-101 dataset is built for the NAS research to alleviate the need for intensive computing resource. NAS-Bench-101 collects the accuracy of architectures tested on CIFAR-10~\cite{krizhevsky2009learning} with Convolutional Neural Networks (CNNs). The NAS-Bench-201 is another benchmark dataset with different search space. In a nutshell, both benchmark datasets choose the cell-based search space, and the whole architecture is built up by stacking different cells which are shown in Fig.~\ref{fig_nas_bench}. The cells in NAS-Bench-101 use vertices to represent operation layers, and edges to represent the connections. Whereas in NAS-Bench-201, the operation layers are presented by edges, which could not use the proposed architecture augmentation directly and need additional transformations to get adjacency matrix and layer type list.

In this section, we will first perform the comparison with the state-of-the-art neural predictors to demonstrate the effectiveness of the proposed HAAP algorithm. Second, HAAP is embedded in an Evolutionary Algorithm (EA)~\cite{yu2010introduction} based NAS search strategy to search for promising architectures on both two benchmark datasets. Third, we will also perform extensive ablation experiments on different classical regression models and on both datasets to show the effect of proposed architecture augmentation method and the utilization of the one-hot encoding strategy.

\subsection{Comparison of Prediction Performance on NAS-Bench-101}
\begin{table*}
	\caption{Comparison with the state-of-the-art neural predictors. We report the results of HAAP on two different $N_{ot}$ for fair comparison.}
	\label{table_state-of-the-art}
	\centering
	\begin{tabular}{c|c|c|c|c}
		\hline
		Algorithms                                   & $N_{ot}$  &  KTau   & MSE     & Regression Model                               \\ \hline \hline
		Peephole~\cite{deng2017peephole}                               & 1K    & $0.4373_{\pm 0.0112}$ & $0.0071_{\pm 0.0005}$  & LSTM                                    \\ \hline
		E2EPP~\cite{sun2019surrogate}                              & 1K    & $0.5705_{\pm 0.0082}$ & $0.0042_{\pm 0.0003}$  & RF                                      \\ \hline
		SSANA~\cite{tang2020semi}                             & 1K    & $0.6541_{\pm 0.0078}$ & $0.0031_{\pm 0.0003}$  & AE + GCN                    \\ \hline
		ReNAS~\cite{xu2021renas}                             & 424   & $0.6574$ & --- & LeNet-5                                 \\ \hline
		NPNAS~\cite{wen2020neural}      &424 & $0.6945$ & --- & GCN \\ \hline \hline
		NPNAS + HA & 424 & $\mathbf{0.7062}$ & --- & GCN \\ \hline
		\multirow{2}{*}{HAAP} 	& 424   & $0.7010_{\pm 0.0022}$ & $0.0024_{\pm 0.0003}$  & \multicolumn{1}{c}{\multirow{2}{*}{RF}}                   \\ 
		& 1K   & $\mathbf{0.7126}_{\pm 0.0024}$ & $\mathbf{0.0023}_{\pm 0.0003}$  & \multicolumn{1}{c}{}                    \\ \hline     
	\end{tabular}
\end{table*}

\begin{figure}
	\centering
	\includegraphics[width=0.8\linewidth]{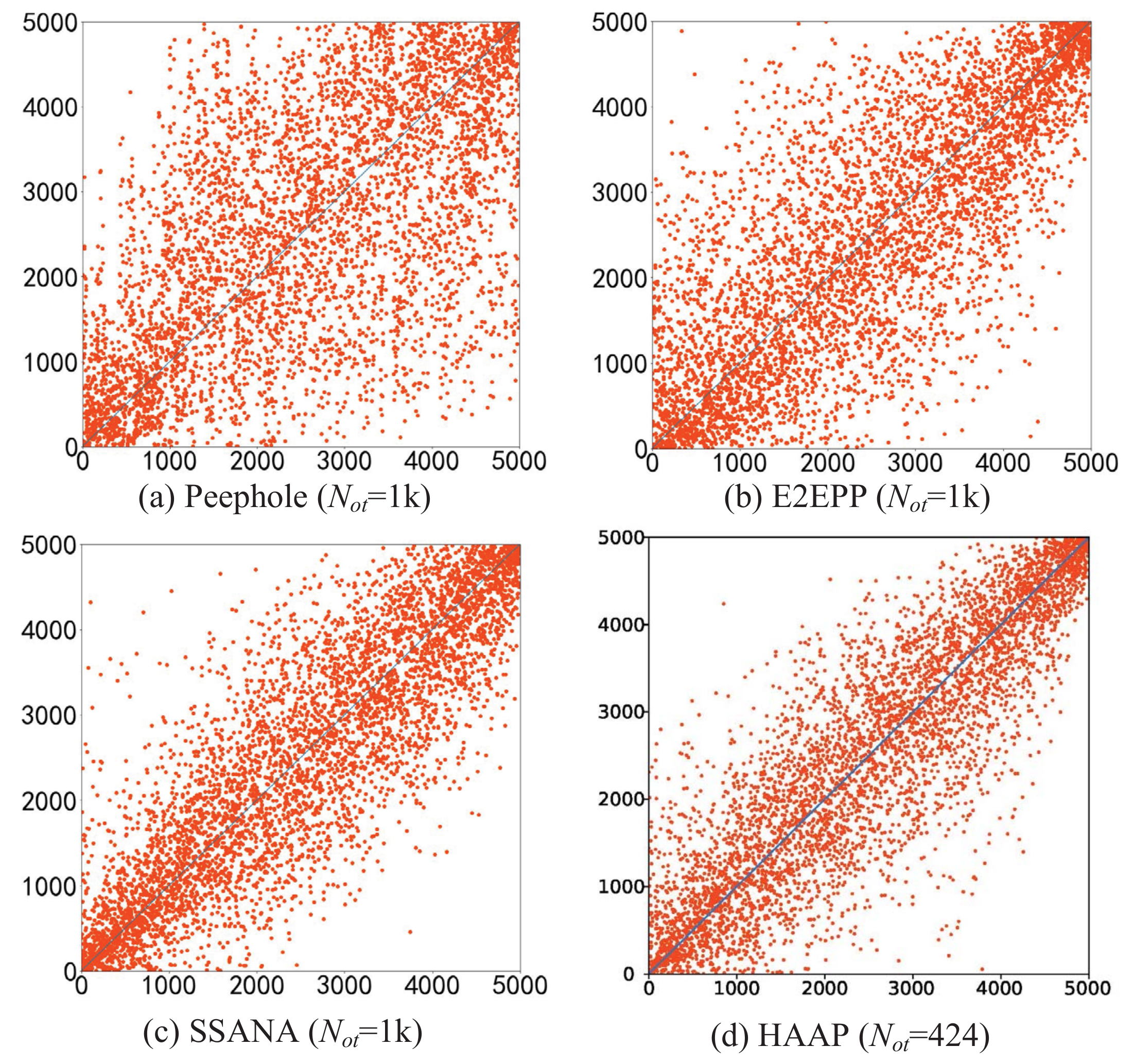}
	\caption{Comparison with state-of-the-art algorithms. The x-axis denotes the true ranking while the y-axis denotes the predicted ranking. Figures (a), (b) and (c) are from~\cite{tang2020semi}.}
	\label{fig_state_comparison}
\end{figure}

The proposed HAAP algorithm is compared with the state-of-the-art algorithms which we have introduced above, i.e., Peephole~\cite{deng2017peephole}, E2EPP~\cite{sun2019surrogate}, SSANA~\cite{tang2020semi}, ReNAS~\cite{xu2021renas} and NPNAS~\cite{wen2020neural}. NAS-Bench-101 is used to providing training and testing architectures.
In addition, two indicators are used to measure the performance of the prediction: Kendall's Tau (KTau)~\cite{sen1968estimates} and MSE. Particularly, the KTau reflects the rank correlation of two ordinal variables, and its value varies in [-1, 1]. The closer this value gets to 1, the more similar the rankings of the two ordinal variables are. The MSE reflects the error between the predicted accuracy values $\hat{\textbf{y}}$ and the ground-truth labels $\textbf{y}$. Please note that both are widely used by the neural predictors of NAS algorithms~\cite{sun2019surrogate,tang2020semi, xu2021renas}.

We report the comparison results in Table~\ref{table_state-of-the-art} where the proposed HAAP algorithm is realized by Random Forest (RF)~\cite{breiman2001random} with fine-tuning. In addition, the symbol ``---" in Table~\ref{table_state-of-the-art} implies there is no result publicly reported in the corresponding literature, and $N_{ot}$ denotes the number of the original training data. The best values are shown in bold for the convenience of observation. For a fair comparison, we report the results of HAAP on two different $N_{ot}$. As can be seen from Table~\ref{table_state-of-the-art}, when $N_{ot}$ is set as 424, HAAP can achieve the largest KTau comparing with ReNAS and NPNAS. In addition, if NPNAS is combined with the proposed Homogeneous Augmentation (HA), the prediction performance can be further improved. With $N_{ot}$ increasing to 1K, HAAP also achieves the best values of KTau and MSE comparing with Peephole, E2EPP and SSANA.

Fig.~\ref{fig_state_comparison} provides the qualitative comparison results, where the x-axis denotes the true ranking while the y-axis denotes the predicted ranking of 5K randomly sampled architectures. If these points are closer to the diagonal, the ranking correlation is stronger, which is equivalent to the values indicated by KTau. Please note that figures (a), (b), (c) are trained with 1K original data whereas $N_{ot}$ in our algorithm is only 424. As can be seen from these figures, the points shown in figure (d) demonstrate the best result, which is achieved by the proposed algorithm. Furthermore, compared with the Neural Networks used by the compared Peephole~\cite{deng2017peephole}, SSANA~\cite{tang2020semi} and ReNAS~\cite{xu2021renas} algorithms (such as Long Short Term Memory (LSTM)~\cite{hochreiter1997long}, Auto-Encoder (AE), Graph Convolutional Network (GCN) and LeNet-5~\cite{lecun1998gradient}), the RF employed by the proposed HAAP algorithm requires few additional tuning of parameters.

\subsection{Searching Architectures on NAS-Bench-101 and NAS-Bench-201}
\begin{table}
	\caption{Classification accuracy rate and ranking of the searched architectures of NAS-Bench-101. 1K annotated architectures randomly selected from NAS-Bench-101 are used to train predictors.}
	\label{table_search_nas_bench_101}
	\begin{center}
		\begin{tabular}{l|l|l}
			\hline
			Algorithm & Accuracy (\%) & Ranking (\%) \\ \hline \hline
			Peephole~\cite{deng2017peephole} & 93.41$\pm$0.34 & 1.64    \\ 
			E2EPP~\cite{sun2019surrogate} & 93.77$\pm$0.13 & 0.15   \\ 
			SSANA~\cite{tang2020semi}  & 94.01$\pm$0.12  &  0.01     \\ \hline
			HAAP & \textbf{94.09}$\pm$0.11 & \textbf{0.004} \\ \hline \hline
			Oracle & 94.23$\pm$0.0 & 0.0005 \\ \hline
		\end{tabular}
	\end{center}
\end{table}

We have referred to the NAS experiment in~\cite{tang2020semi}, and also take an EA based search strategy to search architectures on NAS-Bench-101. In order to demonstrate the effectiveness of the proposed HA, we do not use the state-of-the-art neural predictor when searching. Specifically, the experiment is repeated for 20 times, and we compare the accuracy and ranking of the top-10 architectures selected by different algorithms which can be seen in Table~\ref{table_search_nas_bench_101}. The second column displays the top-1 classification accuracy on CIFAR-10, and the third column displays the ranking of whole architectures in NAS-Bench-101. The last row of Table~\ref{table_search_nas_bench_101} shows the oracle where the performance of all architecture is known. The proposed HAAP obtains the highest accuracy and ranking among the neural predictor algorithms. Specifically, the architectures searched by HAAP are significantly better than that searched by Peephole and E2EPP. Furthermore, HAAP is slightly better than SSANA and is very close to the oracle.

\begin{figure}
	\centering
	\includegraphics[width=1\linewidth]{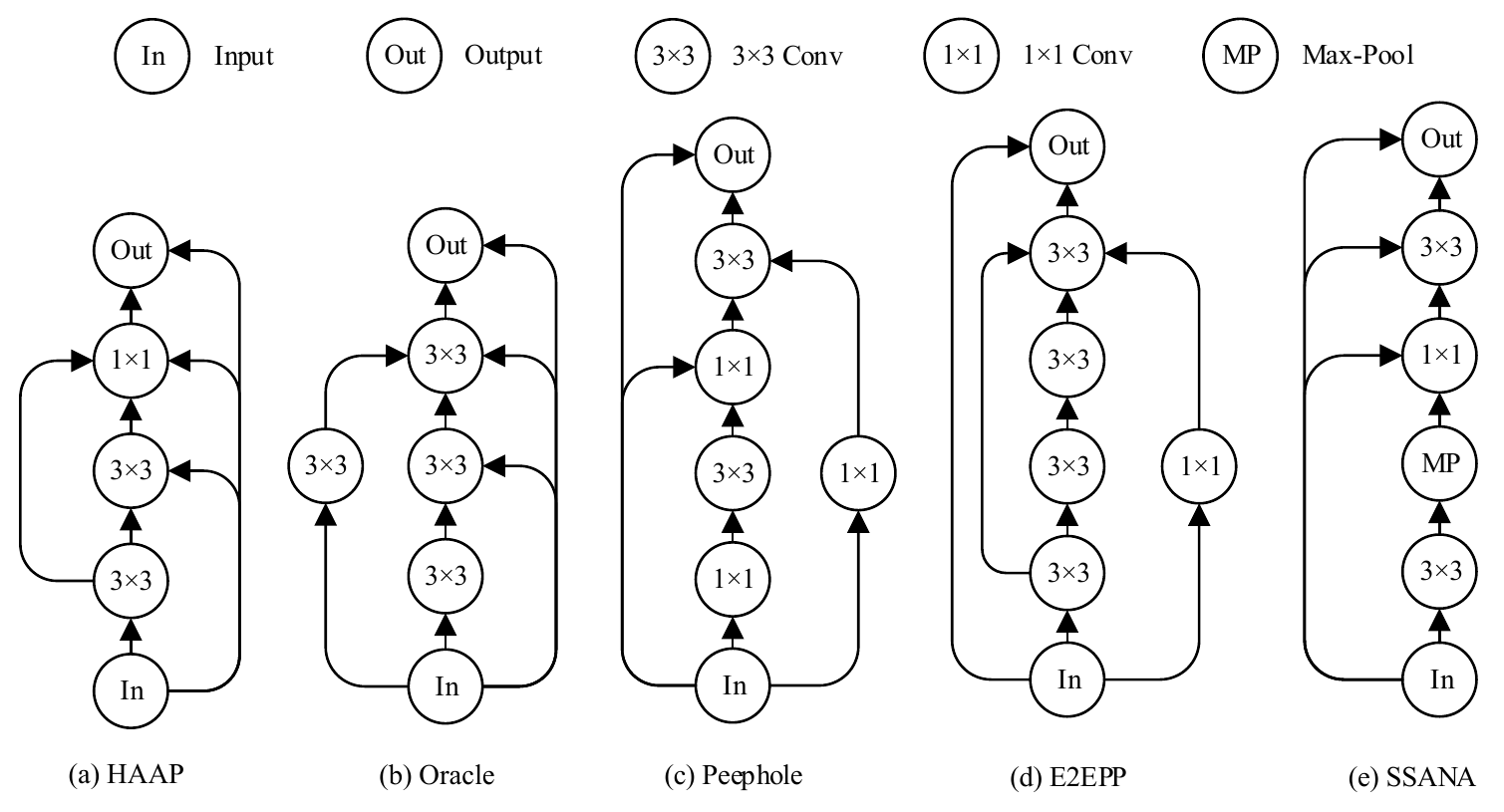}
	\caption{The architectures searched by different neural predictors on NAS-Bench-101. 1K annotated architectures are used to train the neural predictors.}
	\label{fig_searching_result}
\end{figure}

\begin{table}
	\caption{Validation and test accuracy rate on CIFAR-10 of the searched architectures on NAS-Bench-201. We report the mean and std of 10 runs for E2EPP and HAAP.}
	\label{table_sota_nas_bench_201}
	\vspace{-10pt}
	\begin{center}
		\begin{tabular}{l|c|c|c}
			\hline
			Algorithm & Validation & Test & Time \\ \hline \hline
			REA~\cite{real2019regularized} & 91.19$\pm$0.31  & 93.92$\pm$0.30 & 12000  \\ 
			RS~\cite{bergstra2012random} & 90.93$\pm$0.36  & 93.70$\pm$0.36 & 12000\\ 
			REINFORCE~\cite{williams1992simple} & 91.09$\pm$0.37  & 93.85$\pm$0.37  & 12000 \\ 
			BOHB~\cite{falkner2018bohb} & 90.82$\pm$0.53  & 93.61$\pm$0.52 & 12000 \\ \hline
			E2EPP~\cite{sun2019surrogate} & 90.61$\pm$0.89 & 93.39$\pm$0.75 & 6000 \\ \hline
			HAAP   & 91.18$\pm$0.25 & \textbf{94.00$\pm$0.25} & 6000 \\ \hline \hline
			optimal & 91.61 & 94.37 & N/A \\ \hline
		\end{tabular}
	\end{center}
\vspace{-10pt}
\end{table}

We visualize the best architecture searched by different algorithms in Fig.~\ref{fig_searching_result} to give an intuitive comparison. The \textit{In} node is directly connected to the \textit{Out} node in all architectures displayed. This connection can facilitate gradient propagation~\cite{he2016deep, huang2017densely}. There are five operation layers in architectures searched by Peephole and E2EPP, and four in SSANA and oracle. Whereas there are only three operation layers including a 1$\times$1 convolution in HAAP, which can reduce a lot of computing costs.

In order to compare with other state-of-the-art NAS algorithms, we also conduct an experiment searching architectures on NAS-Bench-201. In Table~\ref{table_sota_nas_bench_201}, four algorithms with the top accuracy reported in NAS-Bench-201 and one neural predictor are chosen for comparison. Following the tradition in NAS community, the validation accuracy was used to guide the search during search stage, and the test accuracy was only queried for the searched architecture. To demonstrate the efficiency of HAAP, we stopped HAAP once the time budget of predictor reached the half of the state-of-the-art NAS algorithms (i.e. 6000 seconds). With the half time budget, E2EPP can not exceed any of the competitors. However, HAAP outperforms all the state-of-the-art NAS algorithms, yet with much less search cost.

\subsection{Ablation Study}
\begin{table*}
	\caption{Comparisons on classical regression models on NAS-Bench-101. 424 annotated architectures are used to train the regression models. A and O are short for ``Architecture augmentation" and ``One-hot encoding" separately. The italic and bold denote the best values (highest KTau and lowest MSE) of each regression model.}
	\label{table_classical}
	\begin{center}
		\begin{tabular}{cc|l|lllllllll}
			\hline
			\multirow{2}{*}{A}& \multirow{2}{*}{O} &\multicolumn{1}{c}{\multirow{2}{*}{}} & \multicolumn{9}{c}{Regression Models (Neural Predictor)}                            \\ \cline{4-12} 
			& &\multicolumn{1}{c}{}                  & DT            & LR        & SVR                      & KNN                      & RF            & AdaBoost                 & GBRT                     & Bagging                  & ExtraTree                \\ \hline \hline
			\multirow{2}{*}{\XSolidBrush} & \multirow{2}{*}{\XSolidBrush}     & KTau    & 0.3380                   & 0.3407                   & 0.2738                   & 0.3041                   & 0.5422                   & 0.3624                   & 0.5559                   & 0.4737                   & 0.3410                   \\
			& & MSE     & 0.0048                   & 0.0028                   & 0.0051                   & 0.0028                   & 0.0030                   & 0.0036                   & 0.0031                   & 0.0038                   & 0.0039                   \\ \hline
			\multirow{2}{*}{\XSolidBrush} & \multirow{2}{*}{\Checkmark}     & KTau    & 0.4386                   & 0.5074                   & 0.3257                   & 0.4426                   & 0.5909                   & \textbf{\textit{0.4185}}                   & 0.5779                   & 0.5405                   & 0.4442                   \\
			& & MSE     & 0.0048                   & 7.7862                   & 0.0050                   & \textit{\textbf{0.0027}} & 0.0031                   & 0.0032                   & 0.0030                   & 0.0040                   & 0.0049                   \\ \hline \hline
			\multirow{2}{*}{\Checkmark} & \multirow{2}{*}{\XSolidBrush}     & KTau    & 0.5271                   & 0.3592                   & 0.3256                   & 0.3932                   & 0.6888                   & 0.3158                   & 0.5746                   & 0.6620                   & 0.5154                   \\
			& & MSE     & \textit{\textbf{0.0030}} & 0.0029                   & 0.0046                   & 0.0029                   & 0.0023                   & 0.0028                   & \textit{\textbf{0.0025}} & \textit{\textbf{0.0024}} & 0.0033                   \\ \hline
			\multirow{2}{*}{\Checkmark} & \multirow{2}{*}{\Checkmark}     & KTau    & \textit{\textbf{0.5424}} & \textit{\textbf{0.5456}} & \textit{\textbf{0.3548}} & \textit{\textbf{0.5068}} & \textit{\textbf{0.6991$\star$}} & 0.3309                   & \textbf{\textit{0.6007}}                   & \textit{\textbf{0.6628}} & \textit{\textbf{0.5189}}                   \\
			& & MSE     & 0.0032                   & \textit{\textbf{0.0026}} & \textit{\textbf{0.0045}} & 0.0029                   & \textit{\textbf{0.0022$\star$}} & \textit{\textbf{0.0026}} & \textit{\textbf{0.0025}} & \textit{\textbf{0.0024}} & \textit{\textbf{0.0031}} \\ \hline     
		\end{tabular}
	\end{center}
\vspace{-5pt}
\end{table*}

\begin{table}
	\caption{Results on NAS-Bench-201. The regression model is RF. The bold denotes the best accuracy of each $N_{ot}$ on CIFAR-10.}
	\label{table_nas_bench_201}
	\begin{center}
		\begin{tabular}{cc|l|l|l}
			\hline
			\multirow{2}{*}{A}& \multirow{2}{*}{O} &\multicolumn{1}{c}{\multirow{2}{*}{}} & \multicolumn{2}{c}{$N_{ot}$}                                                              \\ \cline{4-5} 
			& &\multicolumn{1}{c}{}                    & 424        & 1K         \\ \hline \hline
			\multirow{2}{*}{\XSolidBrush} & \multirow{2}{*}{\XSolidBrush}     & KTau      &  $ 0.6919_{\pm 0.0040} $      & $ 0.7754_{\pm 0.0051} $         \\
			& & MSE          & $ 0.0009_{\pm 0.00006} $    & $ 0.0004_{\pm 0.00002} $                      \\ \hline  
			\multirow{2}{*}{\XSolidBrush} & \multirow{2}{*}{\Checkmark}     & KTau    &  $ 0.7156_{\pm 0.0060} $        & $ 0.7647_{\pm 0.0050} $         \\
			& & MSE       & $ 0.0008_{\pm 0.00004} $    & $ 0.0004_{\pm 0.00003} $                      \\ \hline \hline
			\multirow{2}{*}{\Checkmark} & \multirow{2}{*}{\XSolidBrush}    & KTau      &  $ 0.7824_{\pm 0.0014} $      & $ \mathbf{0.8172}_{\pm 0.0051} $    \\
			& & MSE          & $ 0.0005_{\pm 0.00001} $        & $ 0.0004_{\pm 0.00005} $            \\ \hline  
			\multirow{2}{*}{\Checkmark} & \multirow{2}{*}{\Checkmark}     & KTau      &  $\mathbf{0.7899}_{\pm 0.0011}$      & $ 0.8166_{\pm 0.0025} $    \\
			& & MSE          & $ \mathbf{0.0004}_{\pm 0.00005} $        & $ \mathbf{0.0003}_{\pm 0.00001} $            \\ \hline  
		\end{tabular}
	\end{center}
\end{table}

In this subsection, the classical regression models on different cases are compared to show the effectiveness benefiting from the architecture augmentation and the one-hot encoding strategy. They are Decision Tree (DT)~\cite{breiman1984classification}, Linear Regression (LR), Support Vector Regression (SVR)~\cite{chang2011libsvm}, K-Nearest Neighbors (KNN), RF~\cite{breiman2001random}, AdaBoost~\cite{freund1997decision}, Gradient Boosted Regression Tree (GBRT)~\cite{friedman2001greedy}, Bagging~\cite{breiman1996bagging} and ExtraTree~\cite{geurts2006extremely}. Please note that all the models are implemented by Scikit-learn~\cite{pedregosa2011scikit} with the default settings that often give rise to satisfactory performance of the corresponding model in reality.

The results are shown in Table~\ref{table_classical}, where all experiments use the same training data and test data. Note that Table~\ref{table_classical} is split into two parts where the top part uses 424 training data to train the regression model, while the bottom part uses architecture augmented (actually $424\times(7-2)!=50880$ training data) and the test data are all 5K. We only report the result of a single run in the table. Four cases are designed to show the efficiency of HAAP:

\textbf{case 1 (baseline)}: Using only the concatenation of the flattened adjacency matrix and integer vector.

\textbf{case 2 (one-hot)}: One-hot encoding for layer type list $x^{t}$ is used on the basis of case 1.	

\textbf{case 3 (architecture augmentation)}: Using augmented data while other settings are the same as case 1.

\textbf{case 4 (architecture augmentation + one-hot)}: One-hot encoding for $x^{t}$ is used on the basis of case 3.

The comparison of the four cases is shown in Table~\ref{table_classical}. As can be seen from the comparison results, almost all the regression models can benefit from these two designed components. When the architecture augmentation is applied, a great promotion is obtained especially using the architecture augmentation and one-hot encoding strategy collectively. Furthermore, RF achieves the best performance in terms of both KTau and MSE indicators which are marked with $\star$ in Table~\ref{table_classical}. In summary, this group of experiments demonstrates the effectiveness of the proposed architecture augmentation method and utilization of the one-hot encoding strategy. 

In addition, this ablation study is also experienced on NAS-Bench-201. As introduced above, we firstly transform the architectures to the standard DAG form. Table~\ref{table_nas_bench_201} shows the KTau and MSE under four cases and different $N_{ot}$. The values of KTau and MSE in case 1 are much better than that of NAS-Bench-101 as shown in Table~\ref{table_classical}. This demonstrates that the proposed transformation method is effective, and the architectures in NAS-Bench-201 are more recognizable. The cases with architecture augmentation still get the highest KTau and the lowest MSE which is in bold as shown in Table~\ref{table_nas_bench_201}. In addition, under the same setting of $N_{ot}$, the results are better than that of NAS-Bench-101 which can be seen in Table~\ref{table_state-of-the-art}. One important reason is that $N_{l}$ is $8$ in this dataset, and the training data will be expanded by 720 times rather than 120 times in NAS-Bench-101. However, there is no obvious improvement using the one-hot encoding strategy, and this may be caused by the increasing size of adjacency matrix.

\begin{figure}
	\centering
	\includegraphics[width=0.8\linewidth]{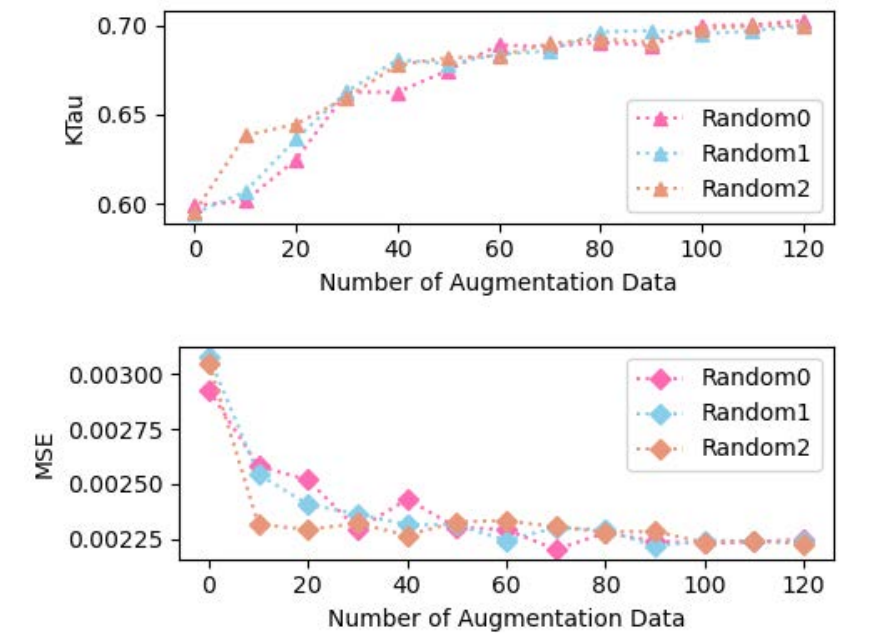}
	\caption{Neural predictor performance results of HAAP w.r.t. different number of architecture augmentation.}
	\label{fig_ablation_number_of_ad}
	\vspace{-5pt}
\end{figure}

Last but not the least, in order to figure out how many augmentation architectures are needed at least training the predictor to achieve the best performance, we conduct an experiment on NAS-Bench-101, and $N_{ot}$ is set to 424. Fig.~\ref{fig_ablation_number_of_ad} shows three separate randomized trials, where architectures are randomly sampled from all the architectures augmented. 
The x-axes in Fig.~\ref{fig_ablation_number_of_ad} is the number of augmentation data from one original training data, and 0 in x-axis denotes no augmentation is used. As can be observed from Fig.~\ref{fig_ablation_number_of_ad}, the trials can obtain the highest KTau and the lowest MSE when all the architectures augmented are used to train the predictor. Because of this, all the architectures augmented are used to train the predictor in all the experiments except this one.

\section{Conclusion}
\label{sec_conclusion}
The goal of this paper is to develop an efficient neural predictor. The goal has been achieved by two core components. Specifically, we have proposed a homogeneous architecture augmentation method, which could construct more training data from the limited original data. In addition, the one-hot encoding strategy is utilized to transform the DNN architectures, enhance the prediction accuracy made by the neural predictors. The experiments have been conducted on NAS-Bench-101 and NAS-Bench-201, and the significant improvements can be observed in both datasets when architecture augmentation is used. The results compared with five state-of-the-art peer neural predictors show that the proposed HAAP algorithm outperforms all the competitors in terms of both KTau and MSE indicators. Furthermore, well-performed architectures can be found on NAS-Bench-101 and NAS-Bench-201 with 94.09\% accuracy (top 0.004\% of the entire search space) and 94.00\% accuracy separately using HAAP. Our future work will focus on encoding additional features of the architecture to improve the performance with the help of more comprehensive description of the architectures. \\

\noindent\textbf{Acknowledgments.} This work was supported by National Natural Science Foundation of China under Grant 61803277 and Science and Technology Innovation Talent Project of Sichuan Province under Grant 2021JDRC0001, and sponsored by CCF-Baidu Open Fund.

{\small
	\bibliographystyle{ieee_fullname}
	\bibliography{mybib.bib}
}

\end{document}